# A Rule-based/BPSO Approach to Produce Low-dimensional Semantic Basis Vectors Set

[a]Atefe Pakzad        [a]Morteza Analoui[1]

[a]School of Computer Engineering, Iran University of Science and Technology, Tehran, Iran.

a_pakzad@comp.iust.ac.ir

analoui@iust.ac.ir

**Abstract**

We intend to generate low-dimensional explicit distributional semantic vectors. In explicit semantic vectors, each dimension corresponds to a word, so word vectors are interpretable. In this research, we propose a new approach to obtain low-dimensional explicit semantic vectors. First, the proposed approach considers the three criteria Word Similarity, Number of Zero, and Word Frequency as features for the words in a corpus. Then, we extract some rules for obtaining the initial basis words using a decision tree that is drawn based on the three features. Second, we propose a binary weighting method based on the Binary Particle Swarm Optimization algorithm that obtains $N_B$ = 1000 context words. We also use a word selection method that provides $N_S$ = 1000 context words. Third, we extract the golden words of the corpus based on the binary weighting method. Then, we add the extracted golden words to the context words that are selected by the word selection method as the golden context words. We use the ukWaC corpus for constructing the word vectors. We use MEN, RG-65, and SimLex-999 test sets to evaluate the word vectors. We report the results compared to a baseline that uses 5k most frequent words in the corpus as context words. The baseline method uses a fixed window to count the co-occurrences. We obtain the word vectors using the 1000 selected context words together with the golden context words. Our approach compared to the Baseline method increases the Spearman correlation coefficient for the MEN, RG-65, and SimLex-999 test sets by 4.66%, 14.73%, and 1.08%, respectively.

## 1. Introduction

For many natural language processing applications, it is important to identify semantic similarities or relatedness between words. Distributional semantic models obtain the word vectors based on the co-occurrence of a target word with the words in a corpus. Distributional semantic models fall into two categories: The first category is the count-based models. The second category is prediction-based models that mainly use neural methods. The count-based models produce explicit word vectors in which each vector component refers to a lexical word. Prediction-based models produce implicit word vectors [1]. The components of implicit word vectors produced by prediction-based models have no lexical equivalents. Prediction-based models provide higher accuracy than count-based models. But unfortunately, the resulting word embeddings are not interpretable.

---

[1] Corresponding Author

Implicit word vectors have been used in many natural language processing tasks, such as sentiment classification [2-4], POS tagging [5], named entity recognition [6], question answering [7], information retrieval [8], and recommendation systems [9-10]. Implicit semantic vectors generated by prediction-based methods have been widely used in sentiment classification tasks. Implicit word vectors do not use the sentiment information of the texts because each dimension of the implicit word vector has no lexical equivalent [2]. So, low dimensional explicit word vectors can help sentiment analysis tasks to use sentiment information of the texts. Also, low dimensional explicit word vectors help other NLP tasks such as recommender systems, question answering, and information retrieval because these vectors are fully interpretable and the text information is reflected in these vectors. In Section 1.1, we describe the construction of semantic word vectors in count-based models.

## 1.1 Distributional semantic vectors

For the construction of explicit word vectors in count-based models, important context words must be identified. Then, the co-occurrence of the target words with the context words is counted and the word-context matrix is obtained. In the word-context matrix, each row corresponds to a target word and each column corresponds to a context word [11]. Generally, in count-based models, a window surrounding the target word is considered to use smaller contexts. For example, when window=5 is considered, it uses 5 words to the left and 5 words to the right of the target word to count the co-occurrence of the target word with the context words. Table 1 shows the co-occurrence vectors for four words which are computed from the Brown corpus. This Table shows only six dimensions of the word-context matrix. Note that a real vector has vastly more dimensions [12].

|             | aardvark | …   | computer | data | pinch | result | sugar | …   |
|-------------|----------|-----|----------|------|-------|--------|-------|-----|
| apricot     | 0        | …   | 0        | 0    | 1     | 0      | 1     |     |
| pineapple   | 0        | …   | 0        | 0    | 1     | 0      | 1     |     |
| digital     | 0        | …   | 2        | 1    | 0     | 1      | 0     |     |
| information | 0        | …   | 1        | 6    | 0     | 4      | 0     |     |

Table 1-co-occurrence vectors of target words [12]

Table 1 shows that the two words *apricot* and *pineapple* are more similar to each other. The words *sugar* and *pinch* occur in their window. Also, the two words *information* and *digital* are more similar to each other. These two words are not similar to the word *apricot*.

After counting the co-occurrences, a high dimensional sparse matrix X is constructed. Each matrix cell $X_{i,j}$ contains the association between the target word $w_i$ and the context word $c_j$. A well-known measure for the association is PMI. The association criterion is very effective in computing the similarity of words. PMI is the log ratio between $P(w,c)$ and $P(w)P(c)$. The PMI is defined as follows [13]:

$$PMI(w,c) = \log \frac{P(w,c)}{P(w)P(c)} \qquad (1)$$

Joint probability $P(w,c)$ determines how often two words co-occur. The product of marginal probability $P(w)P(c)$ informs us that how often we expect the two words to co-occur when they each occur independently. Positive PMI (PPMI) is a more common approach that replaces all negative values by 0 [13]:

$$PPMI(w,c) = \max(PMI(w,c), 0) \qquad (2)$$

So, the PPMI measure should be applied to $X_{i,j}$ that corresponds to the $i_{th}$ target word and the $j_{th}$ context word. The PPMI measure removes the bias of the raw co-occurrence numbers of the matrix X. Each row of the resulting PPMI matrix corresponds to a target word vector. The similarity or relatedness between word pairs is obtained by comparing the semantic vectors of the words.

In this paper, we select $N_B$ important context words using the Binary Particle Swarm Optimization algorithm. In Section 1.2 we explain the general principles of the Particle Swarm Optimization algorithm.

**1.2 Particle swarm optimization**

Kennedy and Eberhart (1995) have developed a heuristic method called PSO [14]. Their idea comes from swarms in nature such as a flock of birds, schools of fish, a swarm of bees, etc. PSO performs its searches by the population of particles. This population of particles is driven by natural swarms using communications based on evolutionary computations. In this algorithm, a particle provides a candidate solution.

The PSO algorithm uses a set of particles flying in the problem space. Flying particles follow the optimal particles and move to a promising area to achieve a global optimum. The velocity of each particle is considered a potential solution. For every time instant, the velocity of each particle changes based on the particles' self-experiences (pbest) and their social experiences (gbest). The components of the PSO algorithm are variables, particles, swarms, and processes. Particles are the candidate solutions and start their fly from random positions in the search space [15].

The PSO algorithm was first developed for continuous problems. Due to the discrete nature of many problems, Kennedy and Eberhart (1997) proposed the discrete version of the PSO [16]. The discrete version of PSO uses discrete binary variables. In the following, we briefly describe the continuous and discrete versions of the PSO algorithm.

*1.2.1 Continuous particle swarm optimization*

Swarms of particles start their fly from random positions in the search space and try to find new solutions in the search space. The position and velocity of the $i_{th}$ particle in the $k_{th}$ iteration are denoted by $x_k^i$ and $v_k^i$, respectively. Therefore, the velocity and position of the $i_{th}$ particle in the $(k + 1)_{th}$ iteration are obtained based on the following equations:

$$x_{k+1}^i = x_k^i + v_{k+1}^i \qquad (3)$$

$$v_{k+1}^i = w \cdot v_k^i + c_1 \cdot r_1 \cdot (p_k^i - x_k^i) + c_2 \cdot r_2 \cdot (p_k^g - x_k^i) \qquad (4)$$

Which $c_1$ and $c_2$ represent cognitive and social components which are arbitrary constants. They can be used to change the weighting between personal and population experience, respectively. Parameters $r_1$ and $r_2$ are random numbers in the range 0 and 1. The best position of the $i_{th}$ particle is determined by $p_k^i$. The best global position in the swarm up to $k_{th}$ iteration is indicated by $p_k^g$ [15].

*1.2.2 Binary particle swarm optimization*

Binary PSO is the implementation of the decision-making for a particle using a discrete decision "true" or " false ". BPSO represents each particle state in the form of binary numbers 0 and 1. Velocity in BPSO is defined based on the changes of probabilities. A specific bit will change the particle solution to zero or one. The sigmoid function is used to map the continuous velocity presented in Equation 5 to [0, 1]. The sigmoid function is shown in Equation 6 [16].

$$V_{k+1}^{ij} = w \cdot V_k^{ij} + c_1 r_1 (p_k^{ij} - X_k^{ij}) + c_2 r_2 (p_k^{gj} - X_k^{ij}) \qquad (5)$$

$$V'^{ij}_k = sig(V^{ij}_{k+1}) = \frac{1}{1 + e^{-V^{ij}_k}} \qquad (6)$$

The superscript i refers to $i_{th}$ particle number. The superscript j refers to $j_{th}$ bit of that particle's velocity. The position of a particle should be updated by equation 7 [17].

$$X^{ij}_{k+1} = \begin{cases} 1 & if\ r^{ij} < sig(V^{ij}_{k+1}) \\ 0 & otherwise \end{cases} \qquad (7)$$

Where $r^{ij}$ is a random number with uniform distribution in the range [0, 1].

In this paper, for the first time, we introduce a new approach to select N informative context words as final basis vectors. First, we consider the most frequent words as candidate context words. Next, we compute WS (word similarity) and NZ (number of zero) criteria for each candidate context word based on implicit word vectors produced by word2vec software. We also use the common WF criterion (word frequency in the corpus) for context word selection. The proposed method extracts some rules to select informative context words based on WF, WS, and NZ criteria using the decision tree. In the next step, we assign a binary weight to each candidate context word using the BPSO algorithm to select $N_B$ context words. We also select $N_S$ context words using the word selection method via distance matrices. Then we extract the golden context words using the binary weighting method. We select context words that their BPSO weights are one for different executions. We add the $N_G$ golden context words to the $N_S$ context words obtained by the word selection method. After selecting the N $=N_S + N_G$ final basis words, we obtain the target word vectors. Then, we evaluate the resulting word vectors in the word similarity task. The results show a significant improvement in the accuracy of the test sets namely, MEN, RG-65, and SimLex-999 datasets.

This paper is structured as follows: Section 2 discusses related works in literature, while section 3 explains the paper's contributions. In Section 4, we describe the methodology and proposed methods to produce semantic basis vectors set. We describe the experimental setups in detail in Section 5. A discussion of the results is given in Section 6. The paper is concluded in Section 7.

## 2. Related Work

In count-based models, word distributions are characterized by high-dimensional and sparse vectors. Because raw co-occurrence counts cause bias, the weighting methods such as tf-idf and PMI are applied to raw vectors. As a result, explicit word vectors are obtained for each target word. Explicit word vectors have high dimensions due to a large number of context words. Therefore, word vectors with high dimensions are mapped to a space of fewer latent dimensions. So, implicit vectors with smaller dimensions appear. This process is called feature extraction because the dimensions of the reduced space are the new features that are extracted from the original data. Thus, implicit vectors are represented with latent semantic space in distributional data. Because explicit word vectors have high dimensions, dimensional reduction methods can be used to produce implicit word vectors. The most common way to create implicit representation is to map a word-context matrix to a reduced latent semantic space with matrix reduction algorithms such as singular value decomposition (SVD), principal component analysis (PCA), and nonnegative matrix factorization (NMF)[18,19].

Recently, predictive or word embedding models infer implicit dense vectors using neural network methods. While count-based models use word-context co-occurrences and generate the explicit word vectors. There are two important issues in semantic space models: 1- Quality of contextual information, and 2- Efficiency. The quality of contextual information and the selection of informative context words is an important issue in semantic space models. Besides, the efficiency factor is very important. Word similarity computations are easier for dense vectors because low-dimensional matrix operations are needed. Cosine similarity is usually used to calculate the similarity between word vectors. Count-based models produce explicit word vectors that have high dimensions, but they can perform effective computations using hashing functions. In the hashing functions, the keys are word-context pairs and their values are non-zero scores. In explicit semantic models, there are many non-existent relations that their value is zero in the word vector.

Reference [20] believes that these non-existent relations can be ignored. As a result, for each word, only context words can be stored that have a non-zero value in the word vector. Most count-based methods transform sparse explicit word vectors to implicit word vectors by dimensional reduction methods. In general, little attention has been paid to the issue of reducing the context words and creating low-dimensional explicit word vectors in count-based models. References [20, 22] use filtering strategies to reduce the dimensions of explicit vectors. They choose the most relevant context words for each word. References [18, 23] describe a count-based model which uses the idea of filtering contexts to reduce non-zero values. The filtering method keeps only the R-relevant context words based on the highest likelihood score in the hash table. Reference [23] say that low dimensional vectors produced by SVD in comparison to sparse matrices that use hash tables for non-zero values are not more efficient computationally. They have shown that vectors with low implicit dimensions do not have more generalizations than explicit word vectors.

Reference [13] show that implicit word vectors generated based on skip-gram and negative sampling implicitly create a word-context matrix. Each component of the word-context matrix is equal to the PMI weight of the word and context pairs. This method is very similar to count-based models that use dimension reduction methods such as SVD to obtain dense implicit word vectors. This shows the similarities between count-based models and prediction-based models. So, good context filters produce low-dimensional explicit word vectors. Low dimensional explicit word vectors can be used in many NLP tasks. Reference [25] in Neuroscience believes that semantic models do not need meaningless latent variables to represent word vectors. So, word co-occurrences should be used instead of reduced latent variables. Implicit word vector dimensions do not have linguistic equivalence. By contrast, explicit word vectors are interpretable because each word vector dimension equals a word.

### 3. Paper contributions

We propose a new approach to construct improved low-dimensional explicit semantic vectors based on a new rule-based selection method and BPSO-based weighting method. Most of the count-based models use the most frequent words in the corpus as context words. We refer to the word frequency in the corpus by the WF criterion. First, we compute a word similarity criterion (WS), the number of zeroes criterion (NZ) and word frequency criterion (WF) for 10k most frequent words in the corpus. We use implicit word vectors obtained by the word2vec method to compute WS and NZ criteria. Second, we extract some rules using a decision tree for finding the most informative context words as initial basis words. We derive some rules using the decision tree of words based on three criteria: WF, WS, and NZ. Based on the extracted rules, informative context words are selected among the 10K most frequently used words in the corpus. These selected context words are called the initial basis words set.

In the next step, we are going to select more important initial basis words as final basis words. We propose a binary weighting method based on the binary particle swarm optimization algorithm for each initial basis word. By applying a binary weight to each initial basis word, weight one means that the initial basis word is useful to select, and weight zero says that the initial basis word is not informative. The proposed binary weighting method helps us to get explicit semantic word vectors in low dimensional. Our method produces a vector space using $N_B$ basis vectors that each basis vector is a natural word. This is the main difference of the proposed binary weighting method compare to the fusion methods such as NMF. Basis vectors in the fusion methods have no direct meaning.

We also use a word selection method based on distance matrices to obtain $N_s$ final basis words. In this method, 5K most frequent words in the corpus are considered as context words. The word selection method selects $N_s$ informative context words via comparison of distance matrices. To the best of our knowledge, the binary weighting method and word selection method are the first approaches that find informative context words from the corpus that works for all vocabulary words. They yield a semantic representation of words satisfying interpretability.

Next, we extract the $N_G$ most informative basis words in the corpus as golden context words applying the binary weighting method. We investigate the effect of adding the $N_G$ golden basis words to the context words obtained by the binary weighting method and the word selection method. The results show that adding $N_G$ golden context words to $N_B$ context words selected by the binary weighting method decreases the Spearman correlation coefficient in test datasets slightly. This result was expected because the $N_B$ context words are produced by the binary weighting method using an optimization algorithm and any change in the selected context words would complicate the optimization process. But adding the $N_G$ golden context words to the $N_S$ context words which are obtained by the word selection method significantly improves the Spearman correlation coefficient in the test sets. This result was also expected because the $N_S$ context words in the word selection method were selected based on the comparison of distance matrices. So by adding $N_G$ golden context words to the $N_S$ context words, we can construct a low dimensional (N=$N_S$+$N_G$) word-context matrix that each row corresponds to an explicit word vector. These low-dimensional explicit semantic vectors can be used for many NLP tasks. The advantage of our approach is that low-dimensional produced word vectors can reflect the information of any particular text using interpretable basis vectors.

## 4. Methodology

In this section, we describe the proposed approach to obtain low-dimensional explicit word vectors. We select N words from the corpus as the final basis words. Each final basis word is equivalent to a basis vector. Therefore, the resulting word vectors are meaningful, so they can reflect the contextual information of the corpus. This approach has several steps, which are described below.

### 4.1. Extracting rules to select initial basis words

In this section, we describe how to select informative basis words of the corpus using the rules derived from the decision tree. In Section 4.1.1, we briefly explain the decision tree and how to extract a set of rules. Section 4.1.2 describes how we extract informative basis words by defining three features WS, NZ, and WF for each word, labeling words, drawing a decision tree, and extracting a set of rules.

*4.1.1. Decision trees and extracting a set of rules*

The decision tree is a hierarchical data structure that can be used for classification and regression applications. In classification applications, algorithms construct decision trees using labeled training samples. An important advantage of the decision tree is its interpretability as well as its ability to be converted into a set of rules. The decision tree is created by a sequence of successive splits [26]. A decision tree has internal decision nodes and final leaves. The output of each leaf is a label that specifies the sample class in the classification problems. Each local area in the input space specifies a class. The boundaries are determined by separators, which are defined using internal decision nodes from root to final leaves. In the decision tree classifier, a path from the root to the leaf determines the conditions for reaching the leaf class label. The conditions for reaching from the root to each leaf can be written using a set of IF-THEN rules. Each rule explains why a class label is selected by the decision tree classifier. The set of extracted rules is called the rule base and it can be used to extract knowledge [26,27].

In Section 4.1.2, we define three features for words to be able to draw their decision tree. Also, we explain in detail how to obtain initial basis words based on the rules extracted from the decision tree classifiers.

*4.1.2. How to select initial basis words set using decision classifier*

Note that C is the candidate set that includes 10K most frequent words in the corpus. We compute two criterions namely word similarity (WS) and the number of zero components (NZ) in the word vector. To calculate the word similarity criterion for a particular word, we compute the similarity of the particular word with the other words in the set C using the implicit word vectors obtained by word2vec software. Also, to obtain the criterion of the number of zero components in a particular word vector, we count the number of zero components of the particular word vector which is obtained by word2vec software. Word vectors obtained by word2vec software have no zero components, so if the absolute value of the vector component is less than 0.01, we assume that the component is zero. We also calculate word frequency in the corpus as a common criterion for obtaining context words. So, for each word in the candidate set C, we compute the word similarity, the number of zero components, and word frequency criteria. Next, we place $\mathcal{M}$ members of set C with the highest word similarity criterion in set S. We place $\mathcal{M}$ members of set C with the highest frequency in set F. Also, we place $\mathcal{M}$ members with the most zero components in the set Z. We define the union of three sets S, F, and Z as triple context words set (Triple set). Also, we place 5k most frequent words of the corpus in set A.

We extract the initial rules for obtaining initial basis words based on the features WS, NZ, WF are given in the following steps:

1) Label the common words of set A and set Triple (set Common) by "1".
2) Label words in set Triple that are not in set A (set IN=Triple-A) by "2".
3) Label the words in set A which are not in set Triple (set OUT = A-Triple) by "3".
4) Let $U_{AT}=A \cup Triple$.
5) Compute WS, NZ, and WF criteria for all words in $U_{AT}$.
6) Draw a decision tree for all words in the set $U_{AT}$.
7) Draw a decision tree for nouns in the set $U_{AT}$.
8) Draw a decision tree for verbs in the set $U_{AT}$.
9) Draw a decision tree for adjectives in the set $U_{AT}$.
10) Draw a decision tree for adverbs in the set $U_{AT}$.
11) Extract the common rules between the 5 decision trees depicted in steps 6 to 10.
12) Normalize the rules obtained in step 11 based on the infinity norm.
13) Select words from candidate set C that satisfies the initial rules in step 12 as IR set.

If the drawn decision trees are examined carefully, we will find some words that have different labels, but have similar features. Therefore, they are considered as noisy samples. The decision tree for all words in the set $U_{AT}$ is shown in Figure 1. Variable x1, x2, and x3 correspond to WS, WF, and NZ criteria, respectively.

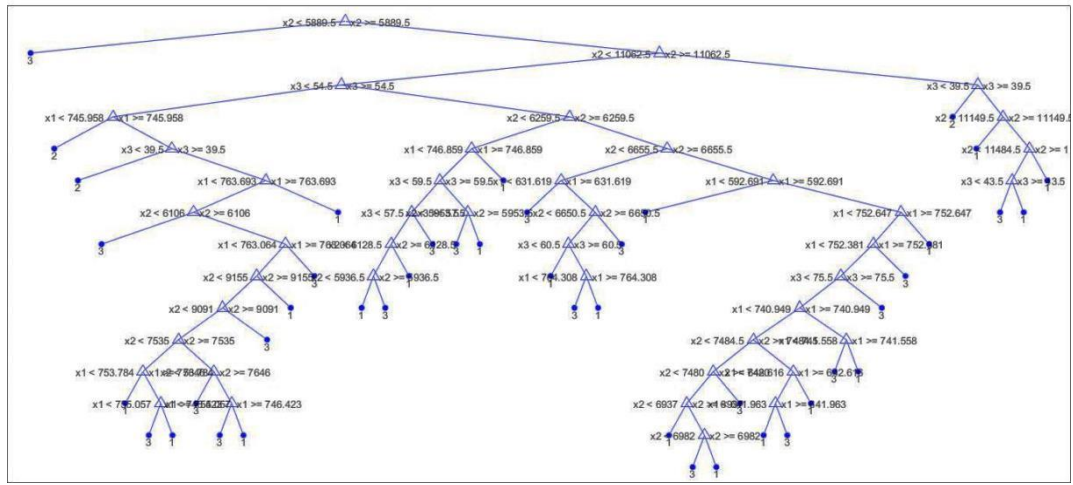

Figure 1- classification tree for all words in the set $U_{AT}$.

We use the classification tree of nouns, verbs, adjectives, and adverbs so that the extracted rules are not biased on a particular POS tag. To find the initial rules:

1) Discard the branches containing the noisy nodes
2) Consider the numbers of nodes in the five decision tree and find the corresponding groups
3) All 5 decision trees start by the WF criterion. So we start the initial rule by the WF criterion.
4) To find the number corresponding to each group, we pay attention to the inequalities. If the inequality is larger, we select the smallest number, and if the inequality is smaller, we select the largest number.

The initial rules extracted from decision trees are as follows.

For each word in set C:

```
IF (WF<5900) {
        Unselect
} ELSE IF (WF≥ 5900) {
        IF (WF ≥ 11063   &&  NZ > 30) {
                Select
        } ELSE IF (WF≥ 8000   &&   WS>700) {
                Select
        } ELSE IF (6600 ≤ WF ≤11063   && WS>763   && NZ>39.5) {
                Select
        } ELSE IF (6600 ≤ WF ≤11063   && WS>643   && NZ>54.5) {
                Select
        } ELSE IF (6600 ≤ WF ≤11063   &&   WS>746   && 39.5 ≤ NZ ≤54.5) {
                Select
        } ELSE IF (5900 ≤ WF ≤6600   &&   WS>763   && NZ ≥ 54.5) {
                Select
        } ELSE IF (WF>7000   &&   NZ ≥ 54.5) {
                Select
        }
}
```

A closer look at the trees reveals that a small WF means that the word is not being selected, and a very large WF criterion is suitable for the selection of a word. But words that have intermediate WF (5900-11063) are needed to satisfy the large WS criterion. We normalize the above initial rules. Each word in set C that satisfies the normalized rules is placed in the IR set. We use $\mathcal{M} = 3000$ to evaluate the Triple context words set. The Triple set contains 4867 words which are noun lemmas, verb lemmas, adjective lemmas, and adverb lemmas. Also, the IR set obtained according to the initial normalized rules contains 3770 words that are nouns, verbs, adjectives, and adverbs. We construct two word-context matrices for the words in vocabulary to evaluate the performance of the sets Triple and IR. We evaluate word vectors on MEN, RG-65, and SimLex-999 test sets. The word-context matrix using set Triple increases the Spearman correlation coefficient in MEN and SimLex-999 datasets that are relatively large test sets. The results show that the initial rules which are extracted based on the three features WS, NZ, and WF are effective. However, the Spearman correlation coefficient in the RG-65 test set is decreased. Due to the reduced accuracy of the RG-65 set, we conclude that some words in the sets IN and OUT are mislabeled.

Examining the selected words of the set IR has the following results:

1) Some words in the sets common and IN cannot satisfy the initial normalized rules. So they are not selected.
2) Some words in set OUT satisfy the initial normalized rules.

These words are likely to be classified incorrectly. Results show that many verb lemmas are selected from the set OUT. About other POS tags (nouns, adjectives, and adverbs), fewer words (about 20 or less) are selected from the set OUT. A large number of words in the set IN can satisfy the initial normalized rules, so we conclude that the selected context words based on the extracted rules are informative. Also, many words in the set OUT cannot satisfy the initial normalized rules. Failure to select a large number of ineffective words in the set OUT is an important factor that improves the word vectors obtained using the set IR. Also, a large number of words in the set common have been selected.

We first try to study the features WS, NZ, and WF in pairs WF-WS, NZ-WS, and NZ-WF. So, for words in the sets IR, Common, IN, and OUT, we plot the word scattering in 2D based on the WF-WS, NZ-WS, and NZ-WF features pairs. Figures 2-4 show the data scattering for the set Common. Figure 2 shows the words scattering based on the WF-WS features pair. Figure 3 shows the words scattering based on the NZ-WF features pair. Also, Figure 4 shows the words scattering based on the NZ-WS pair of features. We examine the plots of the data scattering using the extracted initial rules and the boundaries derived from the decision tree in detail. For each pair WF-WS, NZ-WS, and NZ-WF propose several candidate rules based on the data scattering and the initial rules. We evaluate the context words obtained using each of the candidate rules in each pair of features WS, NZ, and WF. Then, we get the best candidate rule for each features pair. In the next step, we try to combine the rules obtained for each pair of features to obtain candidate rules based on three features WS, NZ, and WF. We evaluate candidate rules based on three features WS, NZ, and WF. Candidate rules based on three features have almost the same Spearman correlation coefficient in the test sets. We choose a rule as the final rule that selects fewer context words. It should be noted that the final rule is more effective than the initial rule. It provides a further improvement in the Spearman correlation coefficient for test sets.

The final normalized rule is as follows:

If (WS>0.6 && NZ>0.24 && WF>0.0015) {

    Select

}

Next, we obtain the context words based on the final normalized rule mentioned above and place them in the set FR. We obtain and evaluate vocabulary word vectors using the selected context words in the set FR. The results show that the Spearman correlation coefficient of word vectors using set FR as context words is higher than those uses set IR as context words. To test the generalizability, we also applied the initial and final normalized rules to the ukWaC4 corpus. Experiment results in similar spearman correlation coefficient compared to the ukWaC1 corpus. It clearly shows that the normalized driven rule is working very well on ukWaC4 despite of the rule is not based on the data of ukWaC4. So, we conclude that the normalized rule is generalized.

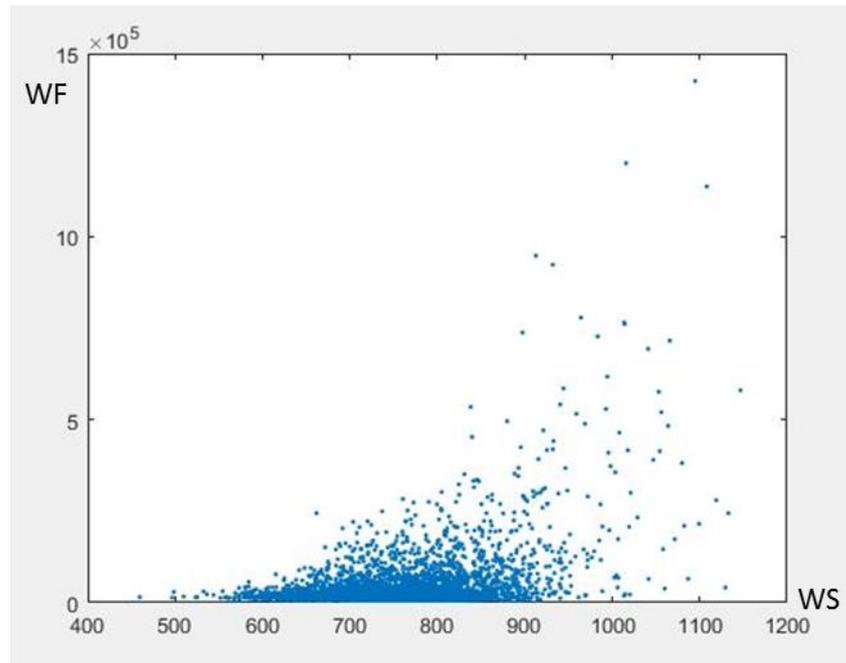

Figure 2- words scattering in 2D based on the WF-WS.

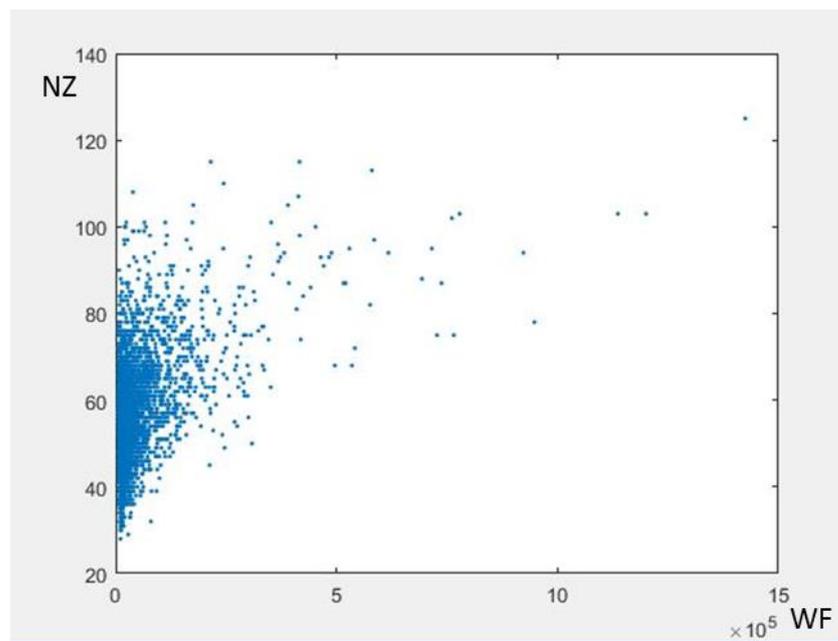

Figure 3- words scattering in 2D based on the NZ-WF

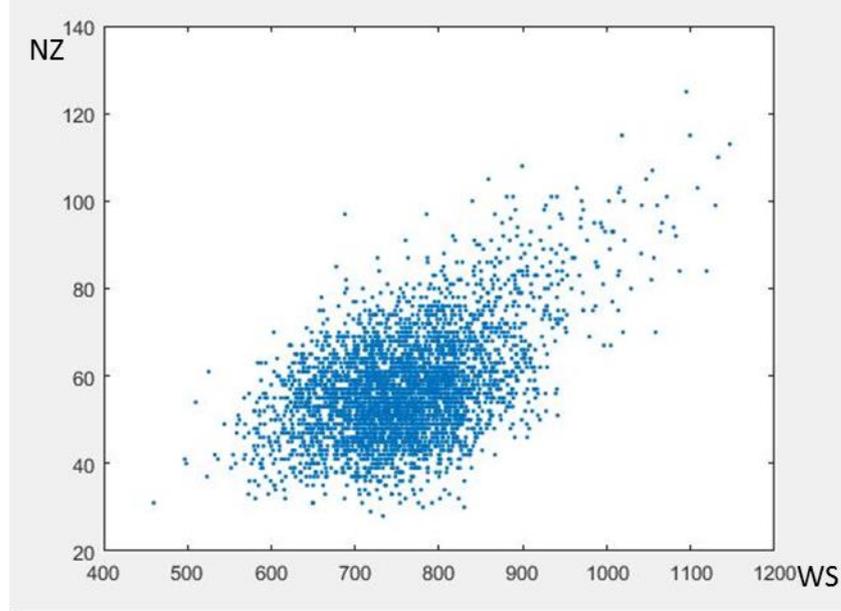

Figure 4- words scattering in 2D based on the NZ-WF

We apply the final rule to the frequent words of the corpus and obtain a relatively smaller set of initial basis words. The set of initial basis words is not small enough. For this reason, in the next section, we introduce an optimal way of context word selection that uses a binary weighting method. Later, we will use the binary weighting method to select golden context words in section 4.3.

### 4.2. BPSO- based weighting method to select $N_B$ context words

In this research, we propose a weighting method based on the BPSO algorithm to select the $N_B$ context words. First, we consider set ℵ as context words set that has $\ell$ words. A word vector is shown in Equation 9.

$$\vec{w} = [v_1, v_2, \ldots, v_\ell] \qquad (9)$$

Then, we compute a binary weight ($bw_i$) for each word $w_i$ using the proposed weighting method.

$$BW = (bw_1, bw_2, \ldots, bw_\ell) \qquad (10)$$

To reduce the number of context words using the optimization algorithm, binary weight "1" is assigned to $N_B$ context words only.

$$\sum_{i=1}^{\ell} bw_i = N_B \qquad (11)$$

The weighting method based on the optimization algorithm tries to obtain the weights of the context words in such a way that the objective function of the problem should be minimized. We define the objective function of the optimization problem as follows:

$$F(BW) = \sum_{i=1,j=1}^{L} \left(\cos(\vec{w}_i, \vec{w}_j) - \cos(\vec{w}'_i, \vec{w}'_j)\right)^2$$

$$= \sum_{i=1,j=1}^{L} \left(\frac{\vec{w}_i \cdot \vec{w}_j}{\|\vec{w}_i\| \|\vec{w}_j\|} - \frac{\vec{w}'_i \cdot \vec{w}'_j}{\|\vec{w}'_i\| \|\vec{w}'_j\|}\right)^2$$

$$= \sum_{\substack{i=1,\\j=1}}^{L} \left(\frac{\sum_{f=1}^{\ell} v_{if} v_{jf}}{\sqrt{\sum_{f=1}^{\ell} v_{if}^2} \sqrt{\sum_{f=1}^{\ell} v_{jf}^2}} - \frac{\sum_{f=1}^{\ell} v_{if} v_{jf} (bw_f)^2}{\sqrt{\sum_{f=1}^{\ell} (v_{if} bw_f)^2} \sqrt{\sum_{f=1}^{\ell} (v_{jf} bw_f)^2}}\right)^2 \quad (12)$$

The training set contains L words. Word vectors ($\vec{w}_i$ and $\vec{w}_j$) use set ℵ as context words. The semantic word vectors $\vec{w}'_i$ and $\vec{w}'_j$ use $N_B$ context words that have binary weight "1".

To solve the problem based on the proposed optimization algorithm, we consider a population with *NP* members. Each member of this population is a particle. Each particle position is equivalent to a weight vector for the context words. That is, each particle position is a vector with $\ell$ components. Each component of the particle position specifies the binary weight of the corresponding context word. The best particle is selected after executing the optimization algorithm. It has the least amount of objective function in comparison to other particles. We consider the best particle position as the final binary weight vector.

The weighting method based on the BPSO algorithm is described in several steps below:

1- Consider NP particles.
2- For each particle, initialize particle position ($x_k^i$) and velocity ($v_k^i$). Particle position should be initialized by binary weights "0" and "1" randomly. Set K=1 which represents the number of repetitions.
3- We apply the objective function in equation (12) to each particle of the population.
4- Do
   a) Particle's best known position ($p_k^i$) should be initialized by $x_k^i$ (means $p_k^i = x_k^i$).
   b) We compute the objective function again for the particles.
   c) The particle's best-known position ($p_k^i$) should be updated by $x_k^i$.
   d) Update the best-known position ($p_k^i$) of each particle and swarm's best-known position ($p_k^g$).
   e) Calculate particle velocity according to the velocity equation (4).
   f) Update particle position according to the position equation (3).
5- While maximum iterations are not attained.
6- Swarm's best-known position ($p_k^g$) is the final weight vector.

*4.2.1. Binary PSO settings*

Parameters $c_1$ and $c_2$ in (4) are constants. They are used to change the weighting between personal and population experience, respectively. In our experiments, cognitive and social components are both set to 0.15. Here, Inertia weight (w) is 0.7. The number of iteration is considered to be 20, which is the stopping criteria. The population size is 30.

**4.3. Finding golden context words**

There are some important words in every textual data that reflect the main information of the text. These golden words can indicate the scope of the text. We call the important words in any textual data as golden words. Golden words also provide information related to sentiment, topic, etc. In this section, we try to get the golden words of the corpus using the PSO-based binary weighting algorithm described in section 4.2. One of the advantages of identifying the golden words of the corpus is that they can be used as the golden basis words to obtain the target word vectors. We do the following steps to find the golden words of the corpus:

1- Obtain the semantic word vectors of the words in the training set using the context words which are selected by the final normalized rule.
2- Apply binary weighting algorithm to word vectors. The binary weighting algorithm must produce a weight vector that has 1000 binary weights "1".
3- Apply the binary weighting algorithm several times (more than 100 times). Then, select the three best solutions that minimize the objective function. So there are three selected binary weight vectors.
4- For each of the three selected weight vectors, find the context words corresponding to the binary weight "1".
5- Find common context words between the three selected solutions. These common words are the golden words of the corpus.

The following selection method introduces another context word selection method named "word selection method".

### 4.4. Word Selection method to select $N_S$ context words

We put the 5K most frequent words in the corpus, which includes nouns, verbs, adjectives, and adverbs in set A. We are going to find $N_s$ informative context words of set A using the word selection method. First, we create a training set that includes h words using the vocabulary words. To find informative context words, for each context word in set A:

1- Create a word-context matrix for the words in the training set using the context words in set A. Each row of the word-context matrix represents a semantic word vector for the corresponding target word. A word vector has 5K components.
2- Create a distance matrix $M_A$ that is $h \times h$ and obtain the Euclidean distance of the word pairs that are in the training set.
3- For all j=1…5K
   a. Remove the j$_{th}$ column from the word-context matrix.
   b. Create a distance matrix $M_{A-\{j\}}$ that is $h \times h$ and obtain the Euclidean distance of the word pairs that are in the training set.
   c. Calculate $M = M_A - M_{A-\{j\}}$.
   d. Calculate the Frobenius norm of Matrix M which determines the level of awareness of the j$_{th}$ context word.
4- Select $N_s$ context words from the set A that have the highest Frobenius norms.

So from the 5K most frequent words in set A, we choose the $\mathbf{N_s}$ informative words as the context words. In step 3.d we use the Frobenius norm to calculate the distance between two matrices ($\mathbf{M = M_A - M_{A-\{j\}}}$). The Frobenius norm of matrix M is obtained as follows [28]:

$$\|M\|_F = (\sum_{i=1}^{h} \sum_{j=1}^{h} (m_{ij})^2)^{1/2} = (\text{trace}(M^T M))^{1/2} \qquad (13)$$

### 4.5. Adding golden context words

We consider the golden words obtained in Section 4.4 as golden context words. We denote the number of golden context words by $N_G$. In the first step, we add the golden context words to the $N_S$ context words which are selected by the word selection method. Then we get the word vectors using $N_S + N_G$ selected context words. We evaluate the obtained word vectors on the test sets. The results show that adding $N_G$ golden context words to the $N_S$ context words which are selected by the word selection method, significantly improves the Spearman correlation coefficient on test sets.

In the next step, we add the golden context words to the $N_B$ context words which are obtained by the binary weighting method. Then we get the word vectors using $N_B + N_G$ selected context words. We evaluate the obtained word vectors on the test sets. The experimental results show that adding $N_G$ golden context words to $N_B$ context words which are obtained by binary weighting method reduces the Spearman correlation coefficient. The accuracy drop is quite expected because the $N_B$ context words are derived based on the optimization algorithm. Therefore, adding golden context words makes the problem of optimality difficult. The results are presented in detail in Section 6. So, by adding the golden words obtained in section 4.3 to the words selected by the word selection method in section 4.4 will significantly improve accuracy. However, the number of context words has been reduced from 5,000 to about 1,080 words.

## 5. Experimental setup

In this section, we describe the details of conducting experiments in this study. We introduce the used corpora in Section 5.1. In the following, the settings for each experiment are described in the corresponding subsection.

### 5.1. Corpus

The ukWaC corpus is a very large corpus for the English language which includes over a billion words. The corpus was created by web crawling. The ukWaC is used as a general-purpose source [29]. The corpus contains the part-of-speech tag and the dependency parsing index. In this article, we perform experiments and parameter adjustments on the first part of the ukWaC namely ukWaC1. Then we examine the ability to generalize experiments on the fourth part of the ukWaC namely ukWaC4.

### 5.2. Construction of semantic word vectors

We define a vocabulary that includes 20K of the most frequent nouns, 10K of the most frequent verbs, 10K of the most frequent adjectives, and 5K of the most frequent adverbs. Then we consider a set of words as the context words set. Then we calculate the co-occurrence number of a target word (vocabulary words) with a context word and place it in the word-context matrix. We use an exponential coefficient $e^{-0.1\alpha}$ to calculate the co-occurrence number of a target word with a context word. Parameter α specifies the distance of the target word from the context word in a sentence. That is, the component $x_{ij}$, which corresponds to the $i_{th}$ target word and the $j_{th}$ context word in the word-context matrix X, is calculated as follows:

$$x_{ij} = \sum_{All\ sentences} e^{-0.1\alpha} \qquad (14)$$

After calculating all the components of the word-context matrix X, the PPMI criterion is applied to each component of the matrix X and the PPMI matrix is obtained. Each row of the PPMI matrix is a semantic vector of a target word. Set A contains the 5K most frequent words (including nouns, verbs, adjectives, and adverbs) in the corpus. The PPMI matrix is called $X_A$ when we use the words in set A as context words.

### 5.3. Finding initial basis words using extracted rules

In Section 4.1, we proposed to extract some rules using the decision tree to obtain the initial basis words. First, we extracted the initial normalized rules. Then, with further analysis, we derived the final normalized rule. The initial and final normalized rules are obtained using the ukWaC1 corpus. To test the generalizability of extracted rules, we derive the initial basis words for the ukWaC4 corpus that satisfy the normalized rules. To begin with, we obtain the three criteria WS, NZ, and WF for the words in candidate set C. Then, using infinity norm, we normalize the 3 criteria of each word. Then each word in set C that can satisfy the initial normalized rule is selected as the initial basis word and we put it in set IR. Any word in set C that can satisfy the final normalized rule is placed in the set FR. Then, we construct word-context matrices $X_{IR}$ and $X_{FR}$ using the initial basis words in sets IR and FR, respectively. We evaluate the word vectors obtained by matrices $X_{IR}$ and $X_{FR}$ on the test sets. These normalized rules have been able to successfully obtain the initial basis words of the ukWaC4 corpus. The results are reported in Section 6.

### 5.4. Context word selection using the binary weighting method

First, we select 2K words in the vocabulary as a training set. Then we construct the word-context matrices $B_A$, $B_{IR}$, and $B_{FR}$ for the words of the training set using the context words in sets A, IR, and FR, respectively. Then we apply the binary weighting algorithm described in section 4.2 to the training set word vectors and obtain $N_B = 1000$ context words with a binary weight "1". The number of weight vector components is equal to the number of context words used to obtain the word vectors. We obtain the vocabulary word vectors again using 1000 selected context words and evaluate them on test sets. The results of the evaluations show that context word selection using the binary weighting method decreases the Spearman correlation coefficient by about 2% on test sets.

### 5.5. Context word selection using the word selection method

We use 5K of the most frequent words (nouns, verbs, adjectives, and adverbs) in the corpus as context words in set A. Also, we build a training set. The training set includes 8K most frequent nouns, 4K most frequent verbs, 4K most frequent adjectives, and 2K most frequent adverbs. We construct the word-context matrix for the words in the training set. Then, we select $N_S$=1000 context words using the word selection method described in Section 4.3. We obtain the vocabulary word vectors using the $N_S$ selected context words. Then, we evaluate the word vectors on test sets. The results show that reducing the 4K context words and using only 1K context words reduces the accuracy by about 2-3%.

### 5.6. Using golden words as golden context words

We find the golden words of ukWaC1 and ukWaC4, using the method described in Section 4.4. Then we choose the common words between the golden words of ukWaC1 and ukWaC4. We consider 20 golden words which are common between the ukWaC1 and ukWaC4 as golden context words. The set SM includes 1000 context words selected by the word selection method. Then we add common golden context words to the set SM. We re-construct vocabulary word vectors using the context words and evaluate them on test sets. The results show that adding golden context words increases the Spearman correlation coefficient significantly.

### 6. Discussion and results

We test the proposed approach for obtaining explicit semantic vectors with small dimensions step by step. The results of the experiments are described in detail in this section. First, we put the 5K most frequent words of the corpus in set A. As a baseline, we obtain word-context matrix $X_{baseline}$ for vocabulary words using words in set A as context words and a window=10. Then we obtain the word-context matrix $X_A$ that uses the exponential coefficient $e^{-0.1\alpha}$ for computing co-occurrence numbers to examine the effect of using the words distances in the sentence. Also, we use set A as context words. The results of evaluating the word-context matrices $X_{baseline}$ and $X_A$ are reported in Table 2. Table 2 shows that the Spearman correlation coefficient of the matrix $X_A$ compared to the matrix $X_{baseline}$ is increased for MEN [30], RG-65 [31], and SimLex-999 [32] datasets by 1.73%, 5.74%, and 1.44%, respectively. The results show that using the distance between the target word and the context word in the sentence for the construction of the word-context matrix has a significant effect on improving word vectors. Note that in this study we obtain word-context matrices using the exponential coefficient $e^{-0.1\alpha}$.

To get the initial basis words, first, we create the set Triple as context words based on the three criteria WS, NZ, and WF as described in section 4.1.2. Then we label the common words in sets Triple and A by "1". Label the words in set IN = Triple-A by "2", which are introduced by WS and NZ criteria as context words. We also label the words in set OUT = A-Triple by "3", which have been frequent words but they are not in the set Triple. Then we draw the decision tree for the labeled words based on the WS, NZ, and WF features. Then, we extract the initial rules. We normalize the initial rules and apply the initial normalized rules to the words in ukWaC1 corpus. Then, we select the words that satisfy the initial normalized rules. The initial context words that are obtained using the initial normalized rules are placed in the set IR. Then, we obtain and evaluate the word-context matrix $X_{IR}$ based on the context words in set IR. As shown in Table 2, the matrix $X_{IR}$ has 1230 dimensions less than matrix $X_A$. Spearman correlation coefficient of matrix $X_{IR}$ rather than matrix $X_A$ is decreased for MEN and SimLex-999 test sets by 0.33% and 1.35%, respectively. In the RG-65 test set, the Spearman correlation coefficient is increased by 0.71%. The reduced accuracy observed in the SimLex-999 test set was due to some incorrect labeling in the Common, IN, and OUT sets. Further investigations are being conducted to address the mislabeling issue.

|  | Matrix $X_{baseline}$ | Matrix $X_A$ | Matrix $X_{IR}$ |
|---|---|---|---|
| Number of Context Words | 5K | 5K | 3770 |
| MEN dataset | 66.89 | 68.62 | 68.29 |
| RG-65 dataset | 56.96 | 62.70 | 63.41 |
| SimLex-999 dataset | 26.22 | 27.66 | 26.31 |

Table 2 Spearman correlation coefficient of word-context matrices $X_{baseline}$, $X_A$, and $X_{IR}$ for ukWaC1 corpus

As described in Section 4.1.2, we plot the word scattering for the sets IR, Common, IN, and OUT based on the WF-WS, NZ-WS, and NZ-WF pairs of features. Then, for each features pair, we find the best rule base that is extracted based on the boundaries. Then, we obtain a final rule based on the three features WS, NZ, and WF using the rules obtained for each feature pair. The rule base for each feature pair and the final rule based on the three features WS, NZ, and WF are shown in Table 3.

| Features | Rule Base |
|---|---|
| Rule Base 1 based on WF and WS | • $WS > 700, WF > 6000$<br>• $600 \leq WS \leq 700, WF > 8000$ |
| Rule Base 2 based on WS and NZ | • $WS > 700, NZ > 40$<br>• $600 \leq WS \leq 700, NZ > 50$ |
| Rule Base 3 based on NZ and WF | • $WF > 8000, NZ > 30$<br>• $6000 \leq WF \leq 8000, NZ > 50$ |
| Final Rule based on WS, NZ, and WF | • $WS > 700, NZ > 40, WF > 8000$ |

Table 3 Rules for features pairs WF-WS, NZ-WS, NZ-WF, and Final rule based on features WS, NZ, and WF.

To evaluate each rule base, we obtain the selected context words that satisfy the rules in the rule base. Then, we construct a word-context matrix using the selected context words based on the rule base. We evaluate the vocabulary word vectors on the test sets. Table 4 reports the evaluation results of each rule base that is listed in Table 3. The results reported in Table 4 are based on normalized rules. Vocabulary word vectors that use selected context words based on Rule Bases 1, 2, and 3 in comparison to vocabulary word vectors that use context words in set A, have similar or even better performance. Semantic word vectors using selected context words that satisfying the final rule increases the Spearman correlation coefficient by 1% for the RG-65 test set, although it has ignored the 1641 context words. The reported results indicate the good performance of the final rule.

| Context Words based on | Rule Base 1 | Rule Base 2 | Rule Base 3 | Final Rule | Set A |
|---|---|---|---|---|---|
| Number of Context Words | 4121 | 3885 | 4141 | 3359 | 5000 |
| MEN dataset | 68.70 | 68.61 | 68.34 | 68.83 | 68.62 |
| RG-65 dataset | 63.51 | 63.11 | 63.61 | 63.69 | 62.70 |
| SimLex-999 dataset | 27.69 | 27.47 | 27.89 | 27.84 | 27.66 |

Table 4 Spearman correlation coefficient of vocabulary word vectors using context words that are satisfying different extracted rules in ukWaC1 corpus.

We select the initial basis words of ukWaC4 corpus using the initial normalized rules and the final normalized rule to evaluate the generalizability of the extracted rules. Next, we construct the word-context matrices $X_{IR}$ and $X_{FR}$ using the selected context words in sets IR and FR. We evaluate vocabulary word vectors on test sets. The results of the evaluations are reported in Table 5. The matrix $X_{FR}$ has 1621 dimensions less than matrix $X_A$. The Spearman correlation coefficient of the matrix $X_{FR}$ compared to the matrix XA is decreased for the MEN, RG-65, and SimLex999 test sets by 0.14%, 0.56%, and 0.14%, respectively. A slight accuracy drop is justifiable because 1621 dimensions of the word-context matrix are reduced. Table 5 shows that the Spearman correlation coefficient in the matrix $X_{IR}$ is much lower than the matrix $X_{FR}$ for the RG-65 dataset. We have already obtained a similar result in the ukWaC1 corpus. The results show the good performance of the final normalized rule for selecting the initial basis words.

| | Matrix $X_A$ | Matrix $X_{FR}$ | Matrix $X_{IR}$ |
|---|---|---|---|
| Number of Context Words | 5K | 3379 | 3697 |
| MEN dataset | 68.42 | 68.56 | 68.38 |
| RG-65 dataset | 61.77 | 61.21 | 58.98 |
| SimLex-999 dataset | 27.76 | 27.62 | 27.37 |

Table 5 Spearman correlation coefficient of word-context matrices $X_A$, $X_{FR}$, and $X_{IR}$ in ukWaC4 corpus.

Next, we apply the binary weighting method on $X_A$ and $X_{FR}$ matrices to obtain $N_B$= 1000 context words. We find the best weight vector obtained for each of the matrices $X_A$ and $X_{FR}$ and put the resulting context words in sets BA and BFR, respectively. Then we construct the word-context matrices $X_{BA}$ and $X_{BFR}$ using the context words in sets BA and BFR, respectively. Then, we evaluate the resulting word-context matrices on the test sets. Table 6 provides the results of evaluating the word-context matrices obtained using the ukWaC1 corpus. Table 7 reports the evaluation results of the word-context matrices obtained using the ukWaC4 corpus. The word-context matrices $X_{BA}$ and $X_{BFR}$ use only 1K context words for constructing the word vectors. According to the results reported in Table 6, the Spearman correlation coefficient of the matrix $X_{BA}$ in comparison to matrix $X_A$ is decreased by 1.26%, 2.55%, and 1.66% for the MEN, RG-65, and SimLex-999 test sets, respectively. Also, the Spearman correlation coefficient in the matrix $X_{BFR}$ in comparison to matrix $X_{FR}$ for MEN and SimLex-999 sets decreases by 2.08% and 0.61%, respectively. The accuracy is increased for the RG-65 test set by 1.23%. The matrix $X_{BFR}$ has 2359 dimensions less than the matrix $X_{FR}$.

|  | Matrix $X_{BA}$ | Matrix $X_A$ | Matrix $X_{BFR}$ | Matrix $X_{FR}$ |
|---|---|---|---|---|
| Number of Context Words | 1K | 5K | 1K | 3359 |
| MEN dataset | 67.36 | 68.62 | 66.75 | 68.83 |
| RG-65 dataset | 60.15 | 62.70 | 64.92 | 63.69 |
| SimLex-999 dataset | 27.00 | 27.66 | 27.23 | 27.84 |

Table 6 Spearman correlation coefficient of word-context matrices $X_{BA}$, $X_A$, $X_{BFR}$, and $X_{FR}$ in ukWaC1 corpus.

In Table 7, the evaluation results of word-context matrices for the ukWaC4 corpus are reported. In the matrix $X_{BA}$ compared to the matrix $X_A$, the Spearman correlation coefficient is decreased by 2.04%, 0.24%, and 2.16% for the MEN, RG-65, and SimLex-999 test sets. Also, the Spearman correlation coefficient in the matrix $X_{BFR}$ compared to the matrix XFR, for the MEN and SimLex-999 datasets are decreased by 1.49% and 0.58%, respectively. Also in the RG-65 set, accuracy is increased by 3.97%. According to the results presented in Tables 6 and 7, by reducing the number of basis words to 1K using the binary weighting method, about 1-2.5% accuracy drop occurs in the test sets. The accuracy drop is justifiable because the word vectors have 1K interpretable dimensions.

|  | Matrix $X_{BA}$ | Matrix $X_A$ | Matrix $X_{BFR}$ | Matrix $X_{FR}$ |
|---|---|---|---|---|
| Number of Context Words | 1K | 5K | 1K | 3379 |
| MEN dataset | 66.38 | 68.42 | 67.07 | 68.56 |
| RG-65 dataset | 61.53 | 61.77 | 65.18 | 61.21 |
| SimLex-999 dataset | 25.60 | 27.76 | 27.04 | 27.62 |

Table 7 Spearman correlation coefficient of word-context matrices $X_{BA}$, $X_A$, $X_{BFR}$, and $X_{FR}$ in ukWaC4 corpus.

In this step, we select $N_S = 1000$ context words from set FR, using the word selection method and place them in the SFR set. Also, we select $N_S = 1000$ words from set A and place them in set SA, using the word selection method. Then, we construct the word-context matrices $X_{SA}$ and $X_{SFR}$ using the context words in sets SA and SFR, respectively. We evaluate the resulting word-context matrices. The results of evaluating the word-context matrices using the ukWaC1 corpus are reported in Table 8. Table 9 shows the results of evaluating the word-context matrices using the ukWaC4 corpus. Table 8 shows that by selecting 1K context words from set A using the word selection method, the Spearman correlation coefficient in vocabulary word vectors for MEN, RG-65, and SimLex-999 sets is decreased by 2.36%, 2.98%, and 0.94%, respectively. Also, by selecting 1K words from the set FR using the word selection method, there is a 1.7% and 2.76% accuracy drop for the MEN and SimLex-999 test sets, respectively. The accuracy of the RG-65 test set is increased by 4.77%.

|  | Matrix $X_{SA}$ | Matrix $X_A$ | Matrix $X_{SFR}$ | Matrix $X_{FR}$ |
|---|---|---|---|---|
| Number of Context Words | 1K | 5K | 1K | 3359 |
| MEN dataset | 66.26 | 68.62 | 67.13 | 68.83 |
| RG-65 dataset | 59.72 | 62.70 | 68.46 | 63.69 |
| SimLex-999 dataset | 26.72 | 27.66 | 25.08 | 27.84 |

Table 8 Spearman correlation coefficient of word-context matrices $X_{SA}$, $X_A$, $X_{SFR}$, and $X_{FR}$ in ukWaC1 corpus.

Table 9 provides the evaluation results of the word-context matrices using the ukWaC4 corpus. Examining the results, we find that in the matrix $X_{SA}$ compared to the matrix $X_A$, the Spearman correlation coefficient is decreased by 1.77%, 4.65%, and 1.84% for the MEN, RG-65, and SimLex-999 test sets, respectively. By decreasing 2379 context words from the set FR, the accuracy of vocabulary word vectors in the MEN, RG-65, and SimLex-999 test sets is increased by 0.39%, 1.62%, and 0.43%, respectively. A closer look at the results in Tables 8 and 9 reveals that applying the word selection method to the initial basis words obtained according to the final normalized rule is more effective. Comparing the results of Tables 6-9, we find that the accuracy drop in the binary weighting method is less than the word selection method. The binary weighting method uses a BPSO-based optimization algorithm. The word selection method uses the comparison of distance matrices to select context words.

|  | Matrix $X_{SA}$ | Matrix $X_A$ | Matrix $X_{SFR}$ | Matrix $X_{FR}$ |
|---|---|---|---|---|
| Number of Context Words | 1K | 5K | 1K | 3379 |
| MEN dataset | 66.65 | 68.42 | 68.95 | 68.56 |
| RG-65 dataset | 57.12 | 61.77 | 62.83 | 61.21 |
| SimLex-999 dataset | 25.92 | 27.76 | 28.05 | 27.62 |

Table 9 Spearman correlation coefficient of word-context matrices $X_{SA}$, $X_A$, $X_{SFR}$, and $X_{FR}$ in ukWaC4 corpus.

Next, we use the binary weighting method to get the golden words of the corpora ukWaC1 and ukWaC4 in the set G. We add golden context words in set G to the context words in set SA for ukWaC1 and ukWaC4. Then, we compute $X_{SA+G}$ for ukWaC1 and ukWaC4. Evaluation results are reported in Table 10. The Spearman correlation coefficient of the matrix $X_{SA+G}$ compared to $X_{SA}$ using ukWaC1 corpus for MEN, RG-65, and SimLex-999 test sets is increased by 5.32%, 9.36%, and 0.2%, respectively. Also, by adding golden context words in set G using ukWaC4 corpus the accuracy is increased for MEN, RG-65, and SimLex-999 test sets by 5.09%, 10.39%, and 1.44%, respectively. The reported results show that adding golden context words improves the Spearman correlation coefficient dramatically for MEN and RG-65 test sets in comparison to both $X_{SA}$ and $X_A$.

|  | Matrix $X_{SA+G}$ | Matrix $X_{SA}$ | Matrix $X_A$ | Matrix $X_{SA+G}$ | Matrix $X_{SA}$ | Matrix $X_A$ |
|---|---|---|---|---|---|---|
| Corpus | ukWaC1 | ukWaC1 | ukWaC1 | ukWaC4 | ukWaC4 | ukWaC4 |
| Number of Context Words | 1085 | 1K | 5K | 1087 | 1K | 5K |
| MEN dataset | 71.58 | 66.26 | 68.62 | 71.74 | 66.65 | 68.42 |
| RG-65 dataset | 69.08 | 59.72 | 62.70 | 67.51 | 57.12 | 61.77 |
| SimLex-999 dataset | 26.92 | 26.72 | 27.66 | 26.73 | 25.29 | 27.76 |

Table 10 Spearman correlation coefficient of word-context matrices $X_{SA+G}$, $X_{SA}$, and $X_A$ in the ukWaC1 and ukWaC4 corpora.

Next, between the golden words of the two corpora ukWaC1 and ukWaC4, we get the common words and put them in the set $G_c$. We add the words in the set $G_c$ as golden context words to the set SA that are obtained using the word selection method. The results obtained on the ukWaC1 corpus are reported in Table 11. By adding golden context words to the set SA, the Spearman correlation coefficient of the vocabulary word vectors for the MEN, RG-65, and SimLex-9999 test sets is increased by 6.01%, 11.97%, and 0.58%, respectively. Also, the Spearman correlation coefficient in the matrix $X_{SA+G_c}$ is significantly improved compared to the matrix $X_A$ for the MEN and RG-65 test sets by 3.74%, 8.99%, respectively.

The results of adding set $G_c$ to the set SA using the ukWaC4 corpus are reported in Table 11. The results presented in Table 11 show a significant increase in the accuracy of the matrices $X_{SA+G_c}$ compared to the matrices $X_{SA}$. By adding the golden context words to the set BA, which is obtained by the binary weighting method, does not increase the Spearman correlation coefficient, but it decreases the Spearman correlation coefficient by 1-3% on the test sets. This result is quite expected because the context words in sets BA and BFR are obtained using the optimization algorithm, and adding new words to the sets destroys the optimality. Lack of optimality reduces Spearman's correlation coefficient. Because the words in set SA are selected by comparing the distance matrices, a significant increase in the Spearman correlation coefficient occurs by adding the golden context words. The Spearman correlation coefficient for $X_{SA+G_c}$ compared to the matrices $X_{SA}$ is increased for MEN, RG-65, and Simlex-999 test sets by 5.6%, 12.34%, and 1.44%, respectively. The results show that it is effective to add golden context words in set $G_c$ to the context word sets that are selected by the word selection method.

|  | Matrix $X_{SA+G_c}$ | Matrix $X_{SA}$ | Matrix $X_A$ | Matrix $X_{SA+G_c}$ | Matrix $X_{SA}$ | Matrix $X_A$ |
|---|---|---|---|---|---|---|
| Corpus | ukWaC1 | ukWaC1 | ukWaC1 | ukWaC4 | ukWaC4 | ukWaC4 |
| Number of Context Words | 1020 | 1K | 5K | 1020 | 1K | 5K |
| MEN dataset | 72.36 | 66.26 | 68.62 | 72.25 | 66.65 | 68.42 |
| RG-65 dataset | 71.69 | 59.72 | 62.70 | 69.46 | 57.12 | 61.77 |
| SimLex-999 dataset | 27.30 | 26.72 | 27.66 | 27.36 | 25.29 | 27.76 |

Table 11 Spearman correlation coefficient of word-context matrices $X_{SA+G_c}$, $X_{SA}$, and $X_A$ in the ukWaC1 and ukWaC4 corpora.

Table 12 shows that the Spearman correlation coefficient of the matrix $X_{SA+G_c}$ compared to the matrix $X_{baseline}$ using ukWaC1 corpus is increased by 5.47%, 14.64%, and 1.03%, for the MEN, RG-65, and SimLex-999 test sets, respectively. Also, the matrix $X_{SFR+G}$ compared to the matrix $X_{baseline}$ increases the accuracy by 4.66%, 14.73%, and 1.08% for the MEN, RG-65, and SimLex-999 datasets, respectively. The best Spearman correlation coefficients for the low-dimensional explicit word vectors using ukWaC1 corpus for the MEN, RG-65, and SimLex-999 datasets are 72.36%, 71.69%, and 27.30%, respectively. Also, the best accuracy in the ukWaC4 corpus for the MEN, RG-65, and SimLex-999 test sets is 72.25%, 69.46%, and 27.36%, respectively.

| | Matrix $X_{baseline}$ | Matrix $X_{SA+G_c}$ | Matrix $X_{SA+G_c}$ |
|---|---|---|---|
| Corpus | ukWaC1 | ukWaC1 | ukWaC4 |
| Number of Context Words | 5K | 1020 | 1020 |
| MEN dataset | 66.89 | 72.36 | 72.25 |
| RG-65 dataset | 56.96 | 71.69 | 69.46 |
| SimLex-999 dataset | 26.22 | 27.30 | 27.36 |

Table 12 Spearman correlation coefficient of word-context matrices $X_{SA+G_c}$, and $X_{baseline}$.

**7-Conclusions**

In this research, we propose an approach to provide low-dimensional explicit semantic word vectors. The proposed approach includes the following steps:

1. We put 5k most frequent words of the corpus in set A. Then we calculate the word-context matrix using the words in set A as context words considering window = 10 and evaluate the resulting semantic word vectors. The resulting word-context matrix is called $X_{baseline}$.
2. We obtain the word-context matrix $X_A$ using the words in set A as context words. We compute matrix components by the exponential coefficient $e^{-0.1\alpha}$ versus a fixed window.
3. We draw a decision tree based on the three criteria WS, NZ, and WF, and extract the initial rules. Then we normalize the initial rules using the infinity norm. Then, by re-examining the boundaries, we infer the final normalized rule for extracting the initial basis words. Then, we extract the initial basis words from the corpus according to the final normalized rule and place them in the set FR.
4. We suggest the binary weighting method based on the BPSO algorithm for extracting context words. Then, we select $N_B = 1000$ context words from set A and set FR using the binary weighting method and put them in sets BA and BFR, respectively.

5- We apply the word selection method to obtain context words from sets A and FR and place the selected context words in sets SA and SFR, respectively.
6- We extract the golden words of the ukWaC1 and ukWaC4 corpora using the binary weighting method and put them in set G. Also, we consider the common words of set G for ukWaC1 and ukWaC4 as golden context words in set $G_C$.
7- We add the golden context words in set $G_C$ to sets SA, and BA. Then we construct the word-context matrices and evaluate the word vectors.

The matrix $X_A$, which uses the exponential coefficient $e^{-0.1\alpha}$, in comparison to matrix $X_{baseline}$ increases the Spearman correlation coefficient for the MEN, RG-65, and SimLex-999 test sets. The word-context matrix using the initial basis words that are selected by the final normalized rule has almost the same accuracy as the matrix $X_A$. However, the dimensions of the resulting word vectors are reduced. Word-context matrices obtained using sets SA, SFR, BA, and BFR as context words in comparison to matrix $X_A$ decrease the accuracy of MEN, RG-65, and SimLex-999 test sets by about 1-3%. The extracted golden context words in set $G_C$ are presented below:

$G_C$ = {Coventry, intellectual, philosophy, contemporary, outcome, oxford, dr, player, dynamic, guess, objective, moment, suggest, station, take, wing, aircraft, prime, announce, several}

We can release the golden words of ukWaC1 and ukWaC4 for interested researchers. Note that the proposed method can be applied to the favorite corpora to select their golden words. We will see a significant increase in the Spearman correlation coefficient for the test sets by adding the gold context words mentioned above to the sets SA obtained by the word selection method. It is noteworthy that the resulting word vectors have a significant increase in the Spearman correlation coefficient in comparison to both matrices $X_{baseline}$ and $X_A$. We used the ukWaC1 corpus to extract the rules. We then applied the normalized rules on the ukWaC4 corpus to test the generalizability of the normalized rules. Significant improvements have been made to both ukWaC1 and ukWaC4 corpora. We proposed an approach that can significantly improve the accuracy of word vectors by extracting 1K words as context words from the corpus. The resulting low-dimensional word vectors (1020 dimensions) are explicit and can reflect information about the text.

In this research, we introduced an approach that uses several steps (applying the final rule, obtaining basis words by the word selection method, obtaining golden words by binary weighting method) to find a small set of basis vectors. Each basis vector corresponds to a natural word and is meaningful. The semantic word vectors obtained by final basis vectors, although have smaller dimensions (1020) compared to the most frequent context words (5K); report a better Spearman correlation coefficient in the word similarity task.

**Funding**

This research did not receive any specific grant from funding agencies in the public, commercial, or not-for-profit sectors.

**Conflicts of interest/Competing interests**

The authors declare that they have no known competing financial interests or personal relationships that could have appeared to influence the work reported in this paper.

**Availability of data and material**

The data and material that support the findings of this study are available from the corresponding author, upon reasonable request.

**Code availability**

All code for data analysis associated with the current submission is available from the corresponding author, upon reasonable request.